\documentclass[letterpaper, 10 pt, conference]{ieeeconf}  % Comment this line out if you need a4paper

\IEEEoverridecommandlockouts                              % This command is only needed if 
\usepackage{siunitx}
\usepackage{amsmath}
\usepackage{amssymb}
\usepackage{amsfonts}
\DeclareMathAlphabet{\mathcal}{OMS}{cmsy}{m}{n}

\usepackage{nicefrac}
\usepackage{soul}
\usepackage{graphics} % for pdf, bitmapped graphics files
\usepackage{epsfig} % for postscript graphics files
\usepackage{mathptmx} % assumes new font selection scheme installed
\usepackage{times} % assumes new font selection scheme installed
\usepackage[colorinlistoftodos,prependcaption,textsize=tiny]{todonotes}
\usepackage{hyperref}

\usepackage{xcolor}% http://ctan.org/pkg/xcolor
\usepackage{colortbl}
\usepackage{tikz,array,collcell}
\usepackage{pifont}
\usepackage{ragged2e}
\usepackage{booktabs}
\usepackage[justification=justified,singlelinecheck=false,font=footnotesize]{caption}
\usepackage{threeparttable}

\usepackage[ruled,vlined]{algorithm2e}

\makeatletter
\patchcmd{\@makecaption}
  {\scshape}
  {}
  {}
  {}
\makeatletter
\patchcmd{\@makecaption}
  {\\}
  {.\ }
  {}
  {}
\makeatother

\title{\LARGE \bf
NeuralSim: Augmenting Differentiable Simulators \\
with Neural Networks
}

\author{
Eric Heiden${}^*{}^{1}$, David Millard${}^*{}^{1}$, Erwin Coumans$^{2}$, Yizhou Sheng$^{1}$, Gaurav S. Sukhatme$^{1,3}$% <-this % stops a space
\thanks{${}^*$Equal contribution}%
\thanks{$^{1}$Department of Computer Science, University of Southern California, Los Angeles, USA
        {\tt\small \{heiden, dmillard, yizhoush, gaurav\}@usc.edu}}%
\thanks{$^{2}$Robotics at Google, Mountain View, USA
        {\tt\small erwincoumans@google.com}}%
\thanks{$^{3}$G.S. Sukhatme holds concurrent appointments as a Professor at USC and as an Amazon Scholar. This paper describes work performed at USC and is not associated with Amazon.}%
\thanks{This work was supported by a Google PhD Fellowship and a NASA Space Technology Research Fellowship, grant number 80NSSC19K1182.}
}

\usepackage{xspace}

\begin{document}

\maketitle
\thispagestyle{empty}
\pagestyle{empty}

%%%%%%%%%%%%%%%%%%%%%%%%%%%%%%%%%%%%%%%%%%%%%%%%%%%%%%%%%%%%%%%%%%%%%%%%%%%%%%%%
\begin{abstract}
    Differentiable simulators provide an avenue for closing the sim-to-real gap by enabling the use of efficient, gradient-based optimization algorithms to find the simulation parameters that best fit the observed sensor readings.
    Nonetheless, these analytical models can only predict the dynamical behavior of systems for which they have been designed.
    In this work, we study the augmentation of a novel differentiable rigid-body physics engine via neural networks that is able to learn nonlinear relationships between dynamic quantities and can thus model effects not accounted for in traditional simulators.
    Such augmentations require less data to train and generalize better compared to entirely data-driven models.
    Through extensive experiments, we demonstrate the ability of our hybrid simulator to learn complex dynamics involving frictional contacts from real data, as well as match known models of viscous friction, and present an approach for automatically discovering useful augmentations.
    We show that, besides benefiting dynamics modeling, inserting neural networks can accelerate model-based control architectures. We observe a ten-fold speed-up when replacing the QP solver inside a model-predictive gait controller for quadruped robots with a neural network, allowing us to significantly improve control delays as we demonstrate in real-hardware experiments.
    We publish code, additional results and videos from our experiments on our project webpage at \url{https://sites.google.com/usc.edu/neuralsim}.
\end{abstract}

%===============================================================================

\section{Introduction}
	
Physics engines enable the accurate prediction of dynamical behavior of systems for which an analytical model has been implemented. Given such models, Bayesian estimation approaches~\cite{ramos2019bayessim} can be used to find the simulation parameters that best fit the real-world observations of the given system. While the estimation of simulation parameters for traditional simulators via black-box optimization or likelihood-free inference algorithms often requires a large number of training data and model evaluations, differentiable simulators allow the use of gradient-based optimizers that can significantly improve the speed of parameter inference approaches~\cite{murthy2021gradsim}.

\begin{figure}
    \centering
    \includegraphics[width=\columnwidth]{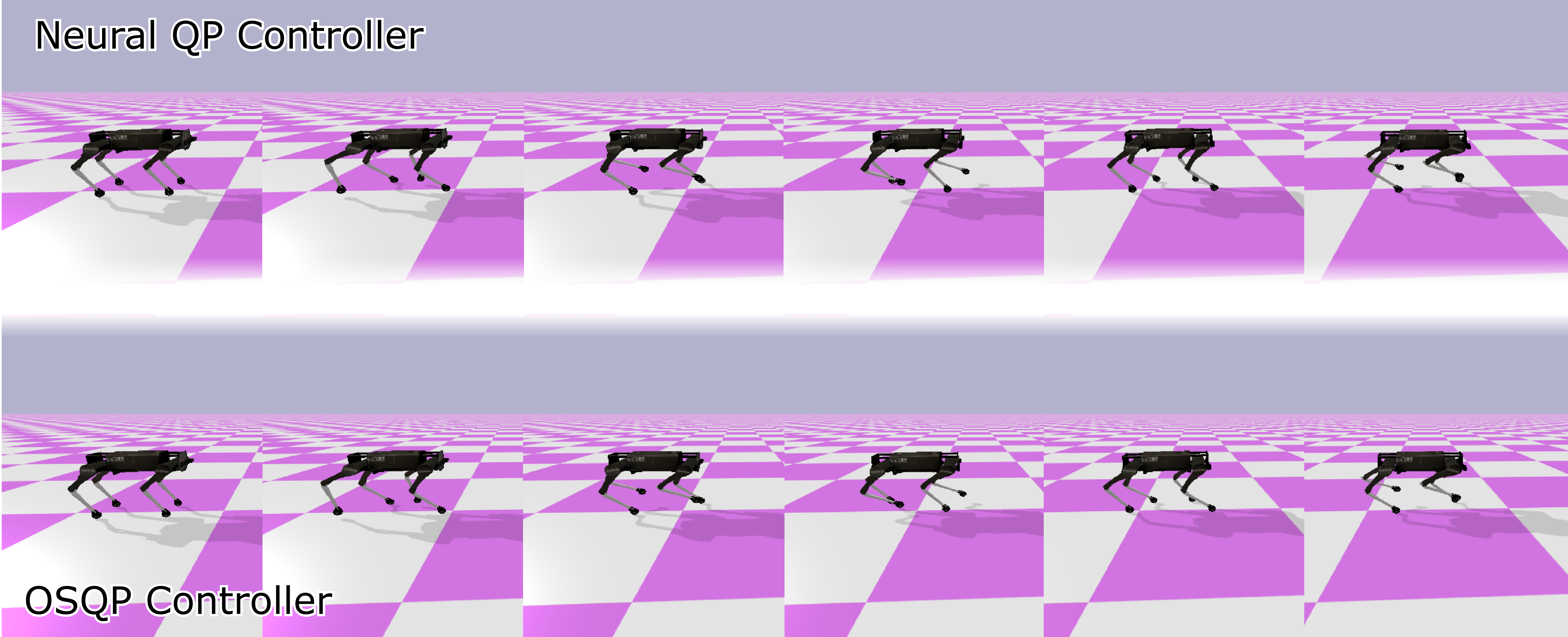}\\\vspace{0.5em}
    \includegraphics[width=\columnwidth]{fig/real_qp_mpc.pdf}
    \caption{\textit{Top:} Snapshots from a few seconds of the locomotion trajectories generated by the QP-based model-predictive controller (first row) and the neural network controller imitating the QP solver (second row) in our differentiable simulator running the Laikago quadruped.
    \textit{Bottom:} model-predictive control using the neural network (third row) and OSQP (fourth row) running on a real Unitree A1 robot.}
    \label{fig:neural-qp}
\vspace{-1em}
\end{figure}

\begin{table*}[t]
    \centering
\begin{threeparttable}
    \begin{tabular}{p{7cm}cccc}
    \toprule
    & \bf Analytical & \bf Data-driven & \bf End-to-end $\nabla$ & \bf Hybrid \\\midrule
    Physics engine~\cite{coumans2013bullet,todorov2012mujoco,lee2018dart} & \checkmark \\
    Residual physics~\cite{hwangbo2019learning,anurag2018hybrid,zeng2019tossingbot,golemo2018neuralaugsim} & \checkmark & \checkmark & & \checkmark \\
    Learned physics~\cite{sanchez2020learning,battaglia2016interaction,li2018learning,jiang2018datacontact} & & \checkmark & \checkmark \\
    Differentiable simulators~\cite{giftthaler2017autodiff,hu2020difftaichi,carpentier2018analytical,peres2018lcp,geilinger2020add,Qiao2020Scalable,murthy2021gradsim} & \checkmark & & \checkmark \\
    Our approach & \checkmark & \checkmark & \checkmark & \checkmark \\
\bottomrule
    \end{tabular}
    \caption{Comparison of dynamics modeling approaches (selected works) along the axes of analytical and data-driven modeling, end-to-end differentiability, and hybrid approaches.}
    \label{tab:comparison}
\end{threeparttable}
\vspace{-1em}
\end{table*}

Nonetheless, a sim-to-real gap remains for most systems we use in the real world. For example, in typical simulations used in robotics, the robot is assumed to be an articulated mechanism with perfectly rigid links, even though they actually bend under heavy loads. Motions are often optimized without accounting for air resistance. In this work, we focus on overcoming such errors due to unmodeled effects by implementing a differentiable rigid-body simulator that is enriched by fine-grained data-driven models.

Instead of learning the error correction between the simulated and the measured states, as is done in residual physics models, our augmentations are embedded inside the physics engine. They depend on physically meaningful quantities and actively influence the dynamics by modulating forces and other state variables in the simulation. Thanks to the end-to-end differentiability of our simulator, we can efficiently train such neural networks via gradient-based optimization.

Our experiments demonstrate that the hybrid simulation approach, where data-driven models augment an analytical physics engine at a fine-grained level, outperforms deep learning baselines in training efficiency and generalizability. The hybrid simulator is able to learn a model for the drag forces for a swimmer mechanism in a viscous medium with orders of magnitude less data compared to a sequence learning model. When unrolled beyond the time steps of the training regime, the augmented simulator significantly outperforms the baselines in prediction accuracy. Such capability is further beneficial to closing the sim-to-real gap. We present results for planar pushing where the contact friction model is enriched by neural networks to closely match the observed motion of an object being pushed in the real world. Leveraging techniques from sparse regression, we can identify which inputs to the data-driven models are the most relevant, giving the opportunity to discover places where to insert such neural augmentations.

Besides reducing the modeling error or learning entirely unmodeled effects, introducing neural networks to the computation pipeline can have computational benefits. In real-robot experiments on a Unitree A1 quadruped we show that when the quadratic programming (QP) solver of a model-predictive gait controller is replaced by a neural network, the inference time can be reduced by an order of magnitude.
    
We release our novel, open-source physics engine, Tiny Differentiable Simulator (TDS)\footnote{Open-source release of TDS: \url{https://github.com/google-research/tiny-differentiable-simulator}}, which implements articulated rigid-body dynamics and contact models, and supports various methods of automatic differentiation to compute gradients w.r.t. any quantity inside the engine.
% Our contributions are three-fold:
% \begin{enumerate}
%     \item We present an architecture for designing differentiable simulators that enables the augmentation of analytical models with data-driven function approximators at any point in the computation graph. This approach generalizes previous residual and entirely data-driven models for learning dynamics.
%     \item Through techniques from sparse regression, we provide an approach to automatically discover where such augmentations need to take place in order to reduce the model error.
%     \item Via extensive experiments, we show how such augmentations drastically reduce the model error. Dynamical effects, such as viscous friction in a fluid that surrounds a swimmer robot, are recovered by finding data-driven models which compensate for the analytical physics engine that previously did not account for such phenomena. We show improved generalization in difficult low-data regime learning problems. Additionally, we present results on real-world datasets and on real robot hardware that demonstrate the sim-to-real capabilities of our approach.
% \end{enumerate}

%===============================================================================

\section{Related Work}
\label{sec:citations}

\begin{figure*}[ht!]
    \centering
    \includegraphics[height=4.5cm]{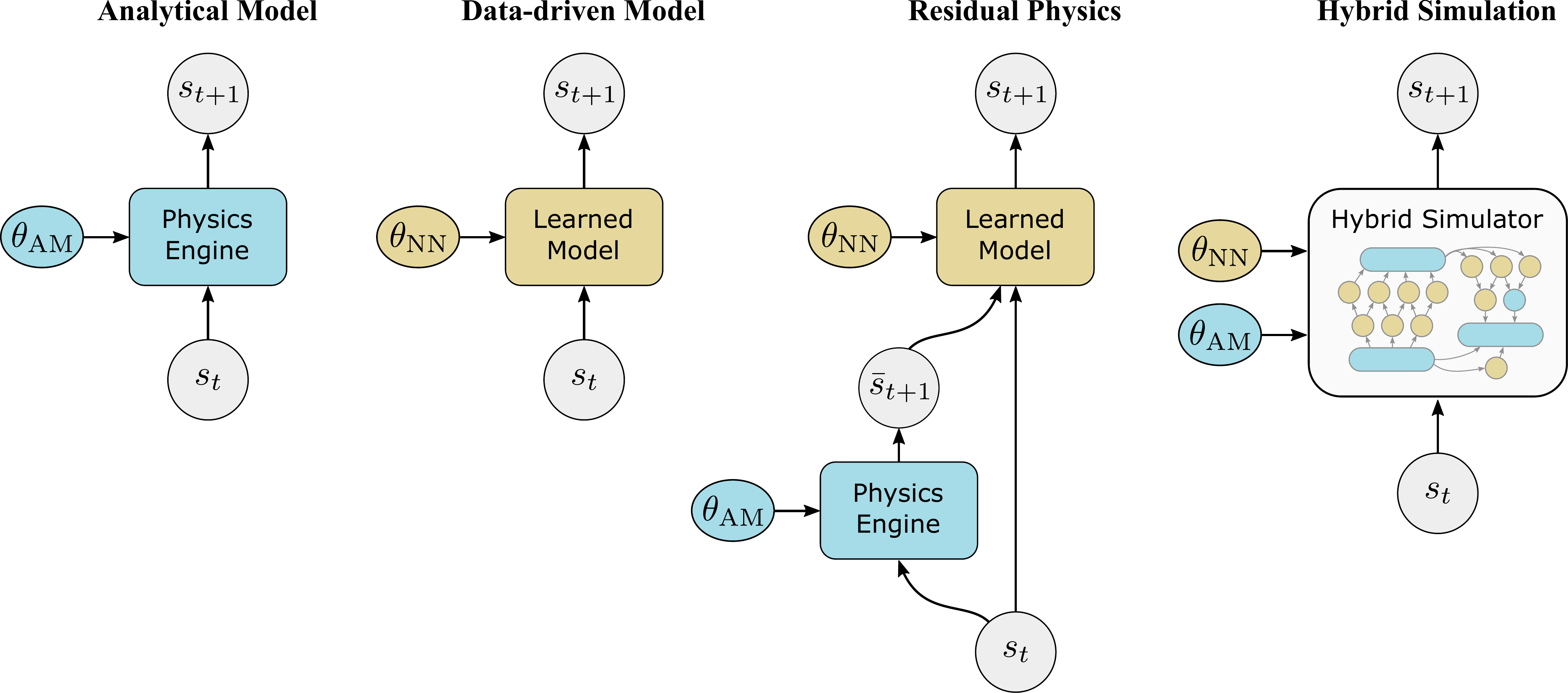}\hfill
    \includegraphics[height=4.5cm]{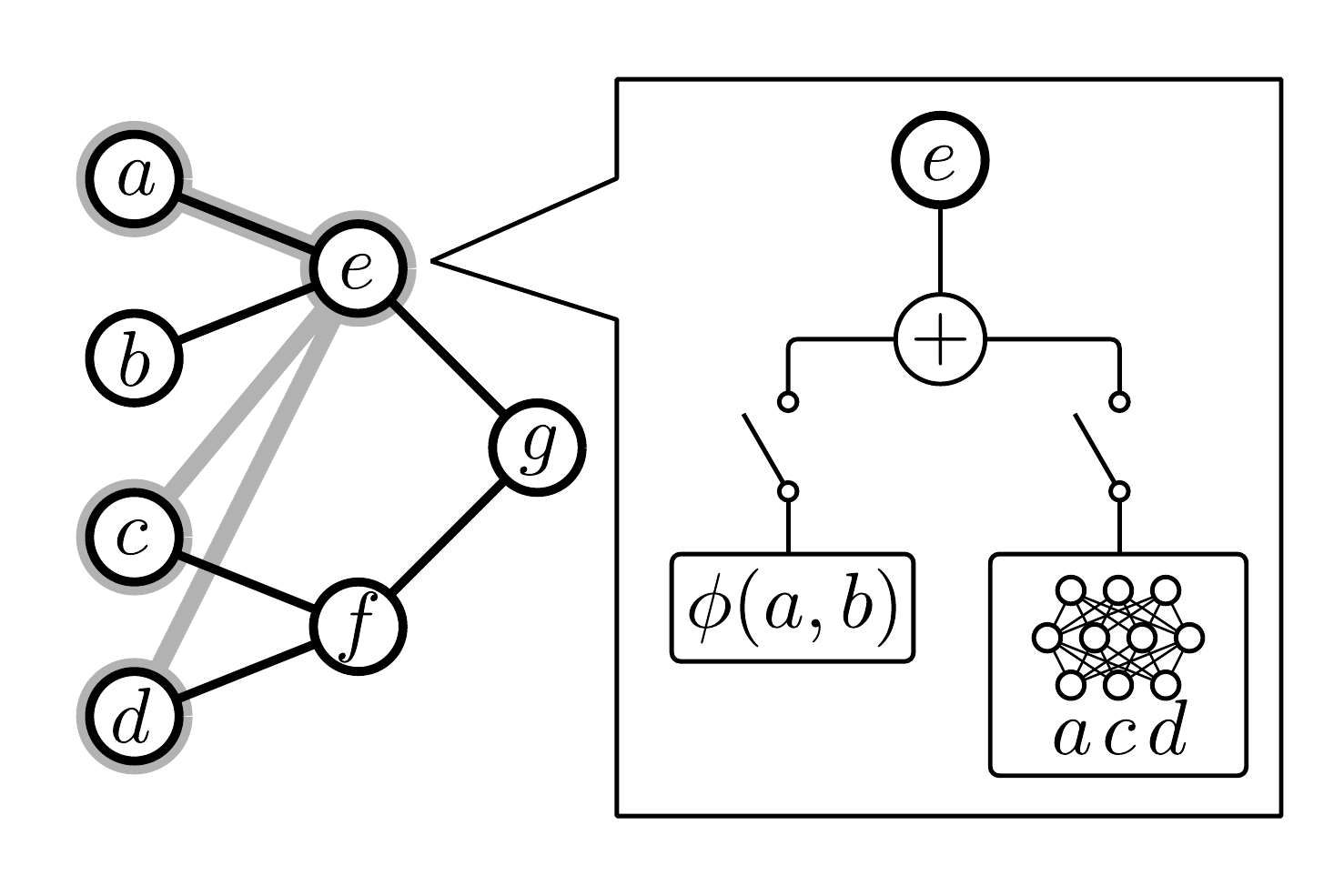}
    \caption{\textit{Left:} comparison of various model architectures (cf.~Anurag et al.~\cite{anurag2018hybrid}). \textit{Right:} augmentation of differentiable simulators with our proposed neural scalar type where variable $e$ becomes a combination of an analytical model $\phi(\cdot,\cdot)$ with inputs $a$ and $b$, and a neural network whose inputs are $a$, $c$, and $d$.}
    \label{fig:approaches}
\end{figure*}

Various methods have been proposed to learn system dynamics from time series data of real systems (see \autoref{tab:comparison} and \autoref{fig:approaches} left). Early works include Locally Weighted Regression~\cite{schaal1994lwr} and Forward Models~\cite{moore1991fm}.  Modern ``intuitive physics'' models often use deep graph neural networks to discover constraints between particles or bodies~(\cite{battaglia2016interaction, xu2019physics, he2019physics, raissi2018physics, mrowca2018physics, chen2018neural, li2018learning, sanchezgonzalez2020learning}). Physics-based machine learning approaches introduce a physics-informed inductive bias to the problem of learning models for dynamical systems from data~\cite{chen2018ode,long2018pde,raissi2019physics,lutter2019delan,greydanus2019hnn,sutanto20rnea}.
We propose a general-purpose hybrid simulation approach that combines analytical models of dynamical systems with data-driven residual models that learn parts of the dynamics unaccounted for by the analytical simulation models.

Originating from traditional physics engines (examples include~\cite{coumans2013bullet,todorov2012mujoco,lee2018dart}), differentiable simulators have been introduced that leverage automatic, symbolic or implicit differentiation to calculate parameter gradients through the analytical Newton-Euler equations and contact models for rigid-body dynamics~\cite{giftthaler2017autodiff,carpentier2018analytical,peres2018lcp, koolen2019rbd-julia, heiden2019ids,  heiden2019real2sim,geilinger2020add,lelidec2021diffsim,lutter2020differentiable}. Simulations for other physical processes, such as light propagation~\cite{NimierDavidVicini2019Mitsuba2, heiden2020lidar} and continuum mechanics~\cite{hu2019chainqueen,liang2019differentiable,hu2020difftaichi,Qiao2020Scalable,murthy2021gradsim} have been made differentiable as well.

Residual physics models~\cite{zeng2019tossingbot,anurag2018hybrid,hwangbo2019learning,golemo2018neuralaugsim} augment physics engines with learned models to reduce the sim-to-real gap. Most of them introduce residual learning applied from the outside to the predicted state of the physics engine, while we propose a more fine-grained approach, similar to~\cite{hwangbo2019learning}, where data-driven models are introduced only at some parts in the simulator. While in~\cite{hwangbo2019learning} the network for actuator dynamics is trained through supervised learning, our end-to-end differentiable model allows backpropagation of gradients from high-level states to any part of the computation graph, including neural network weights, so that these parameters can be optimized efficiently, for example from end-effector trajectories.

\begin{figure}
    \centering
    \includegraphics[width=0.7\columnwidth]{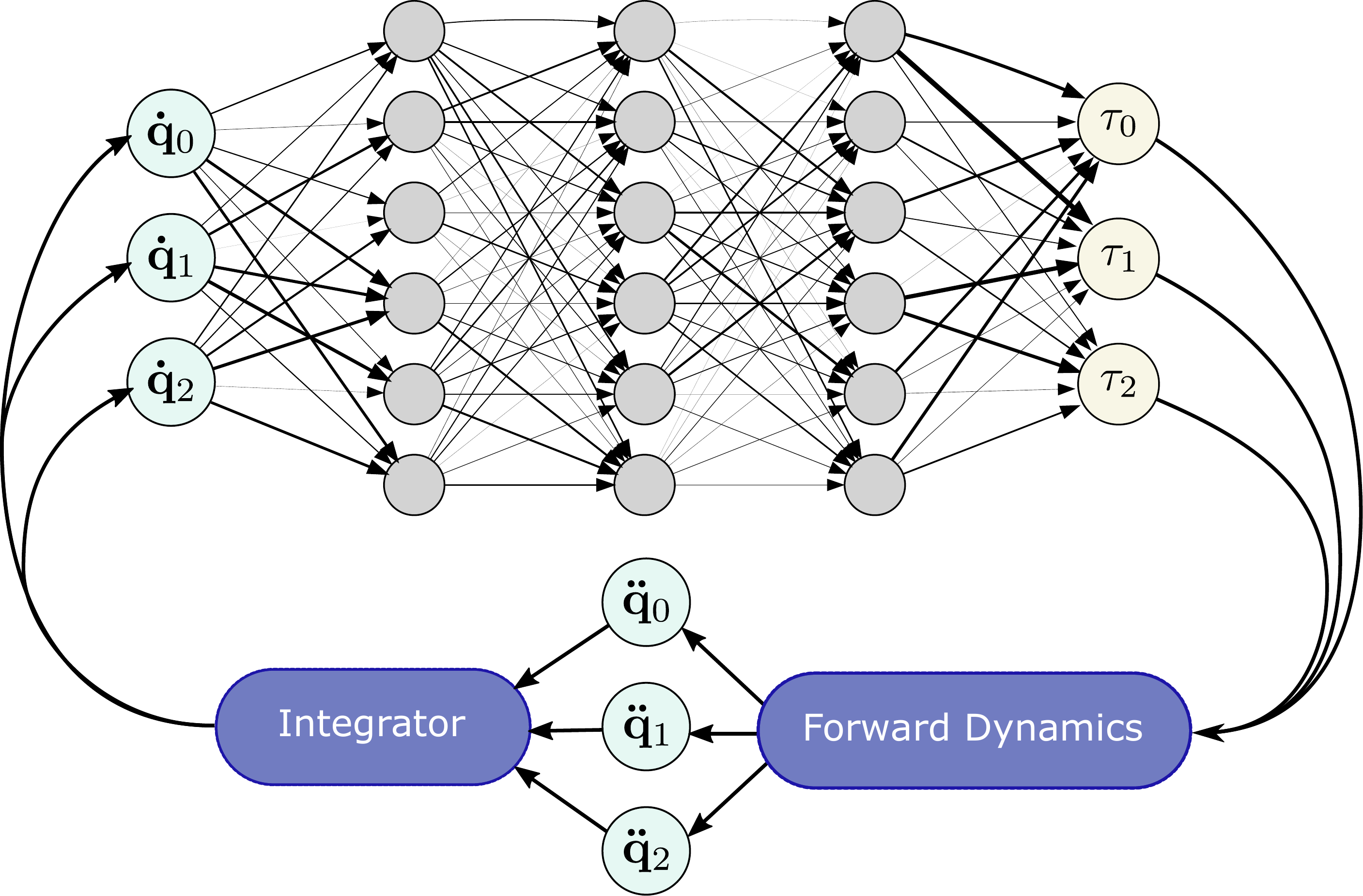}
    \caption{Exemplar architecture of our hybrid simulator where a neural network learns passive forces $\tau$ given joint velocities $\mathbf{\dot{q}}$ (MLP biases and joint positions $\mathbf{q}$ are not shown). }
    \label{fig:golf}
\vspace{-0.25in}
\end{figure}
	
%===============================================================================

\section{Approach}
\label{sec:approach}

Our simulator implements the Articulated Body Algorithm (ABA)~\cite{featherstone2007rbda} to compute the forward dynamics (FD) for articulated rigid-body mechanisms. Given joint positions $\mathbf{q}$, velocities $\mathbf{\dot{q}}$, torques $\tau$ in generalized  coordinates, and external forces $\mathbf{f}_\text{ext}$, ABA calculates the joint accelerations $\mathbf{\ddot{q}}$. We use semi-implicit Euler integration to advance the system dynamics in time. We support two types of contact models: an impulse-level solver that uses a form of the Projected Gauss Seidel algorithm~\cite{stewart2000implicit,horak2019similarities} to solve a nonlinear complementarity problem (NCP) to compute contact impulses for multiple simultaneous points of contact, and a penalty-based contact model that computes joint forces individually per contact point via the Hunt-Crossley~\cite{hunt1975coefficient} nonlinear spring-damper system. In addition to the contact normal force, the NCP-based model resolves lateral friction following Coulomb's law, whereas the spring-based model implements a nonlinear friction force model following Brown et al.~\cite{brown2018friction}.

Each link $i$ in a rigid-body system has a mass value $m_i$, an inertia tensor $\mathbf{I}_i \in \mathbb{R}^{6 \times 6}$, and a center of mass $\mathbf{h}_i \in \mathbb{R}^{6 \times 6}$. Each corresponding $n_d$ degree of freedom joint $i$ between link $i$ and its parent $\lambda(i)$ has a spring stiffness coefficient $k_i$, a damping coefficient $d_i$, a transformation from its axis to its parent ${}^{\lambda(i)}\mathbf{X}_i \in \mathbb{SE}(3)$, and a motion subspace matrix $\mathbf{S}_i \in \mathbb{R}^{6 \times {n_d}}$. 

For problems involving $n_c$ possible contacts, each possible pair $j$ of contacts has a friction coefficient $\mu_j \in \mathbb{R}$ describing the shape of the Coulomb friction cone. Additionally, to maintain contact constraints during dynamics integration, we use Baumgarte stabilization~\cite{baumgarte1972stabilization} with parameters $\alpha_j, \beta_j \in \mathbb{R}$. To soften the shape of the contact NCP, we use Constraint Force Mixing (CFM), a regularization technique with parameters $\mathbf{R}_j \in \mathbb{R}^{n_c}$.
$$
\theta_{AM} = \{m_i, \mathbf{I}_i, \mathbf{h}_i, {}^{\lambda(i)}\mathbf{X}_i, \mathbf{S}_i\}_i \; \cup \; \{\mu_j, \alpha_j, \beta_j, \mathbf{R}_j\}_j
$$
are the \emph{analytical model parameters} of the system, which have an understandable and physically relevant effect on the computed dynamics. Any of the parameters in $\theta_{AM}$ may be measurable with certainty a priori, or can be optimized as part of a system identification problem according to some data-driven loss function.

With these parameters, we view the dynamics of the system as
\begin{align}
\label{eqn:augmented-manipulator-equation}
\mathbf{\tau} = \mathbf{H}_{\theta_{AM}}(\mathbf{q})\mathbf{\ddot{q}} + \mathbf{C}_{\theta_{AM}}(\mathbf{q}, \mathbf{\dot{q}}) + \mathbf{g}_{\theta_{AM}}(\mathbf{q})
\end{align}
where $\mathbf{H}_{\theta_{AM}}$ computes the generalized inertia matrix of the system, $\mathbf{C}_{\theta_{AM}}$ computes nonlinear Coriolis and centrifugal effects, and $\mathbf{g}_{\theta_{AM}}$ describes passive forces and forces due to gravity.

%===============================================================================

\subsection{Hybrid Simulation}
\label{sec:hybrid-sim}

Even if the best-fitting analytical model parameters have been determined, rigid-body dynamics alone often does not exactly predict the motion of mechanisms in the real world. To address this, we propose a technique for hybrid simulation that leverages differentiable physics models and neural networks to allow the efficient reduction of the simulation-to-reality gap. By enabling any part of the simulation to be replaced or augmented by neural networks with parameters $\theta_{NN}$, we can learn unmodeled effects from data which the analytical models do not account for. 

Our physics engine allows any variable participating in the simulation to be augmented by neural networks that accept input connections from any other variable. In the simulation code, such \emph{neural scalars} (\autoref{fig:approaches} right) are assigned a unique name, so that in a separate experiment code a ``neural blueprint'' is defined that declares the neural network architectures and sets the network weights.
	
%===============================================================================

\subsection{Overcoming Local Minima}
\label{sec:basin-hopping}

\begin{figure}
    \centering
    \includegraphics[width=0.9\columnwidth]{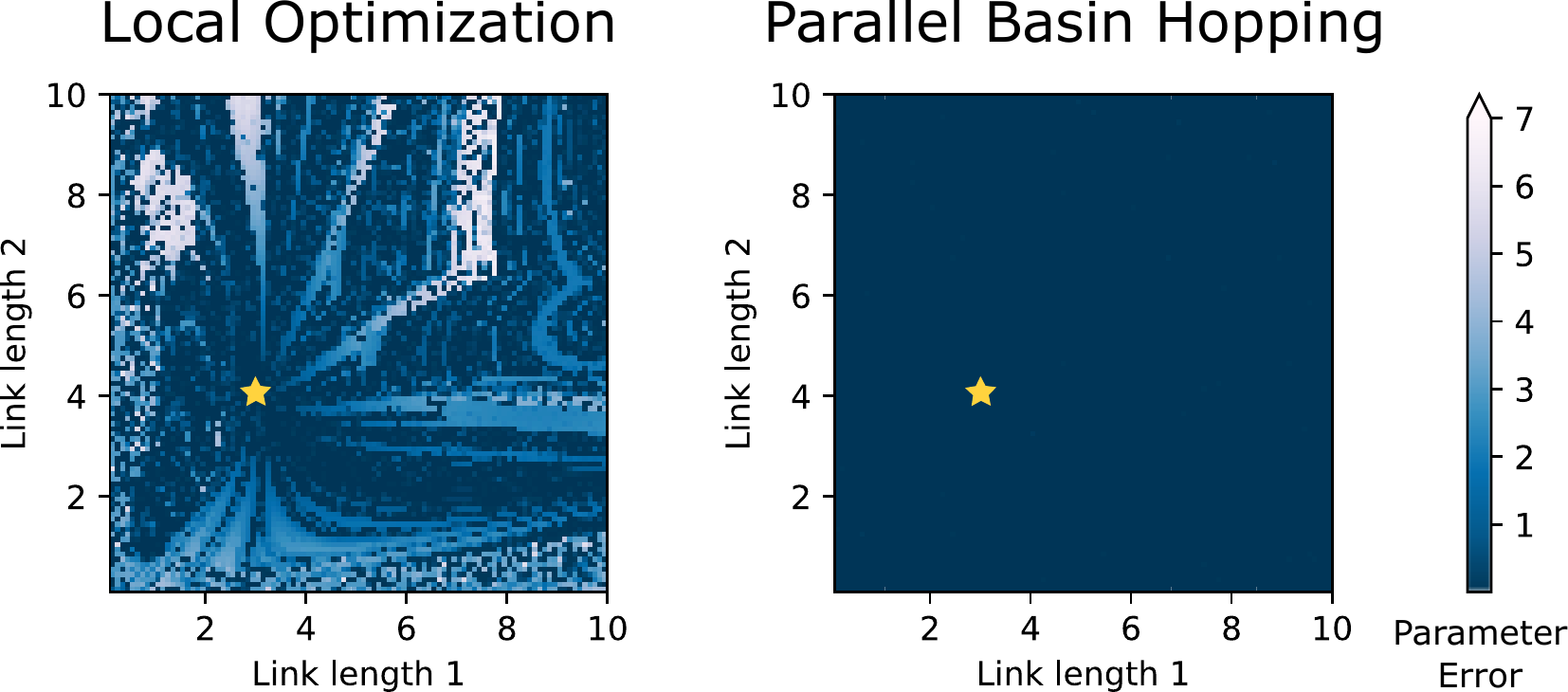}
    % \hfill
    % \includegraphics[width=0.35\columnwidth]{fig/ibm_pendulum_estimation.pdf}
    \caption{
    % \textit{Left:} 
    Comparison of system identification results from local optimization (left) and the parallel basin hopping strategy (right), where the grid cells indicate the initial parameters from which the optimization was started. The link lengths of a double pendulum are the two simulation parameters to be estimated from joint position trajectories. The true parameters are 3 and 4 (indicated by a star), and the colors indicate the $\ell2$ distance between the ground truth and the parameters obtained after optimization (darker shade indicates lower error).
    % \textit{Right:} After system identification of a real double pendulum~\cite{asseman2018learning}, the sim-to-real gap is strongly reduced.
    }
    \label{fig:local-vs-pbh}
\vspace{-0.25in}
\end{figure}

For problems involving nonlinear or discontinuous systems, such as rigid-body mechanisms undergoing contact, our strategy yields highly nonlinear loss landscape which is fraught with local minima. As such, we employ global optimization strategies, such as the parallel basin hopping strategy~\cite{mccarty2018parallel} and population-based techniques from the Pagmo library~\cite{Biscani2020}. These global approaches run a gradient-based optimizer, such as L-BFGS in our case, locally in each individual thread. After each evolution of a population of local optimizers, the best solutions are combined and mutated so that in the next evolution each optimizer starts from a newly generated parameter vector. As can be seen in \autoref{fig:local-vs-pbh}, a system identification problem as basic as estimating the link lengths of a double pendulum from a trajectory of joint positions, results in a complex loss landscape where a purely local optimizer often only finds suboptimal parameters (left). Parallel Basin Hopping (right), on the other hand, restarts multiple local optimizers with an initial value around the currently best found solution, thereby overcoming poor local minima in the loss landscape.

\subsection{Implementation Details}
\label{sec:implementation}

We implement our physics engine \emph{Tiny Differentiable Simulator} (TDS) in C++ while keeping the computations agnostic to the implementation of linear algebra structures and operations.
% To not require third-party libraries, we provide our own collection of linear algebra functions and data structures needed for the rigid-body dynamics computations. 
Via template meta-programming, various math libraries, such as Eigen~\cite{eigenweb} and Enoki~\cite{Enoki} are supported without requiring changes to the experiment code. Mechanisms can be imported into TDS from Universal Robot Description Format (URDF) files.

To compute gradients, we currently support the following third-party automatic differentiation (AD) frameworks: the ``Jet'' implementation from the Ceres solver~\cite{ceres-solver} that evaluates the partial derivatives for multiple parameters in forward mode, as well as the libraries Stan Math~\cite{carpenter2015stan}, CppAD~\cite{cppadweb}, and CppADCodeGen~\cite{cppadcodegen} that support both forward- and reverse-mode AD. We supply operators for taking gradients through conditionals such that these constructs can be traced by the tape recording mechanism in CppAD. We further exploit code generation not only to compile the gradient pass but to additionally speed up the regular simulation and loss evaluation by compiling models to non-allocating C code.

\begin{figure}
    \centering
    \includegraphics[width=0.85\columnwidth]{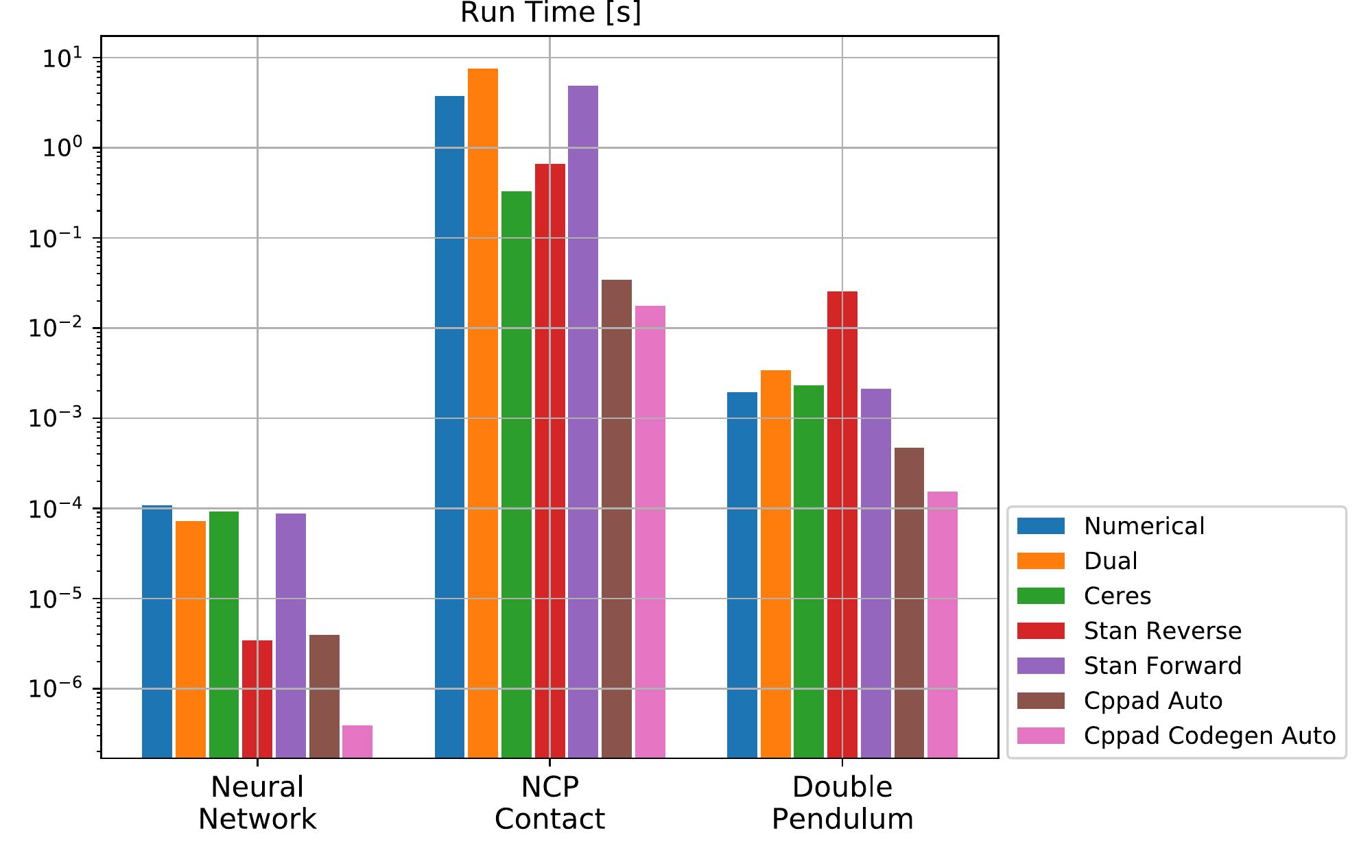}
    \caption{Runtime comparison of finite differences (blue) against the various automatic differentiation frameworks supported by our physics engine. Note the log scale for the time (in seconds).}
    \label{fig:adbench}
\vspace{-0.25in}
\end{figure}

In our benchmark shown in \autoref{fig:adbench}, we compare the gradient computation runtimes (excluding compilation time for CppADCodeGen) on evaluating a 175-parameter neural network, simulating 5000 steps of a 5-link pendulum falling on the ground via the NCP contact model, and simulating 500 steps of a double pendulum without contact. Similar to Giftthaler et al.~\cite{giftthaler2017autodiff}, we observe orders of magnitude in speed-up by leveraging CppADCodeGen to precompile the gradient pass of the cost function that rolls out the system with the current parameters and computes the loss.

%===============================================================================

\section{Experimental Results}
\label{sec:result}

% \subsection{Learning Air Resistance in Golfing}
% \label{sec:exp-golf}

% \begin{figure}
%     \centering
%     \includegraphics[width=0.7\columnwidth,trim=1.5cm 0 0 0]{fig/golf_trajectory_air25_wy=0_19.pdf}
%     \caption{Trajectories of a golf ball in vacuum and air (at 25 ${}^\circ$C temperature). The analytical rigid-body dynamics model of our simulator does not account for air resistance due to the air's density and viscosity. Augmented with a learned model, it can accurately predict the passive forces necessary to match the reference trajectories, generalizing over various golf ball impacts.}
%     \label{fig:golf-traj}
% \end{figure}

% To demonstrate how differentiable simulators can benefit from neural networks augmenting the implemented dynamics equations, we equip a rigid-body physics engine with a learned model that accounts for drag. Such phenomenon has nonlinear effects on the dynamics of objects moving through a medium, such as air, that has nonzero viscosity and density. Our hybrid simulator equipped with a three-layer feed-forward neural network (\autoref{fig:golf}) learns the passive forces necessary to correct the analytical physics engine that models the golf ball's motion in vacuum. Trained on 20 different golf ball pushes (forces), the hybrid simulator generalizes well to other initial conditions (cf. \autoref{fig:golf-traj}).

\subsection{Friction Dynamics in Planar Pushing}
\label{sec:exp-pushing}

Combining a differentiable physics engine with learned, data-driven models can help reduce the simulation-to-reality gap.
Based on the Push Dataset released by the MCube lab~\cite{yu2016more}, we demonstrate how an analytical contact model can be augmented by neural networks to predict complex frictional contact dynamics. 

\begin{figure}
    \centering
    \includegraphics[width=0.55\columnwidth,trim=3em 0 0 3em,clip]{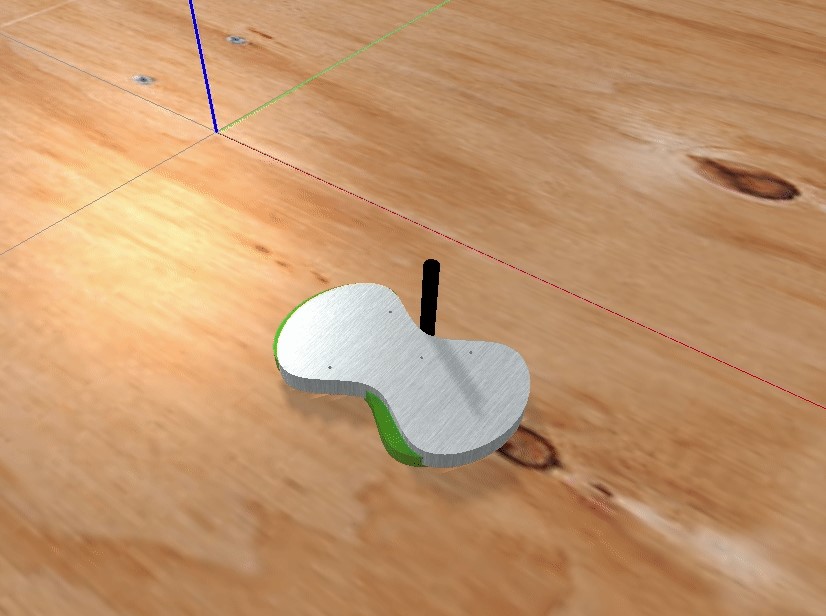}\hfill
    \includegraphics[width=0.40\columnwidth, trim=0 0 2em 0]{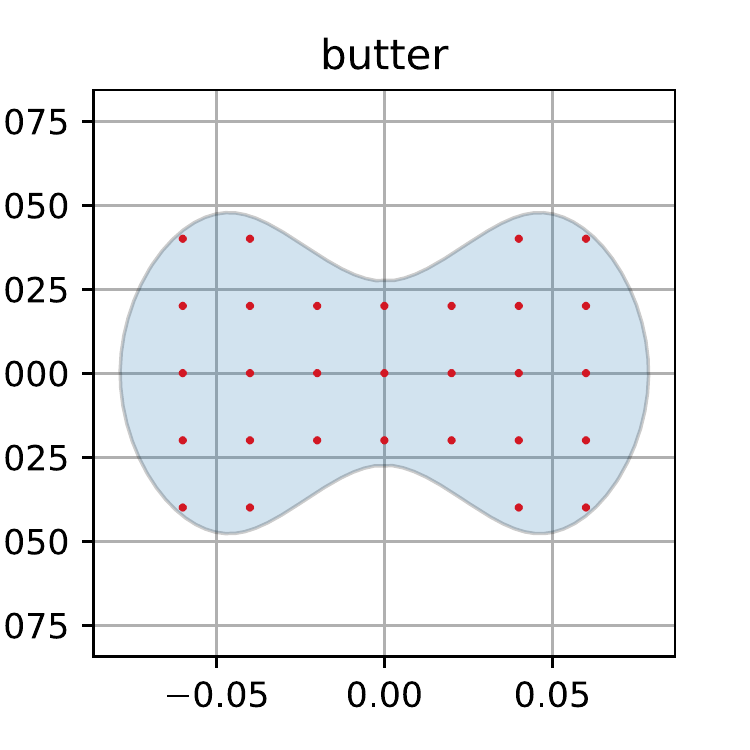}
    \caption{\textit{Left:} example setup for the push experiment (\autoref{sec:exp-pushing}) where the rounded object is pushed by the black cylindrical tip at a constant velocity on a plywood surface. Without tuning the analytical physics engine, a deviation between the pose of the object in the real world (green) and in the simulator (metallic) becomes apparent.
    \textit{Right:} predefined contact points (red dots) for one of the shapes in the push dataset.}
    \label{fig:push-butter-plywood}
\end{figure}

The dataset contains a variety of planar shapes that are pushed by a robot arm equipped with a cylindrical tip on different types of surface materials. Numerous trajectories of planar pushes with different velocities and accelerations have been recorded, from which we select trajectories with constant velocity. We model the contact between the tip and the object via the NCP-based contact solver (\autoref{sec:approach}). For the ground, we use a nonlinear spring-damper contact model that computes the contact forces at predefined contact points for each of the shapes. We pre-processed each shape to add contact points along a uniform grid with a resolution of \SI{2}{\cm} (\autoref{fig:push-butter-plywood} right). Since the NCP-based contact model involves solving a nonlinear complementarity problem, we found the force-level contact model easier to augment by neural networks in this scenario since the normal and friction forces are computed independently for each contact point.

\begin{figure}
    \centering
    \includegraphics[width=0.6\columnwidth,trim=3em 3em 3em 3em]{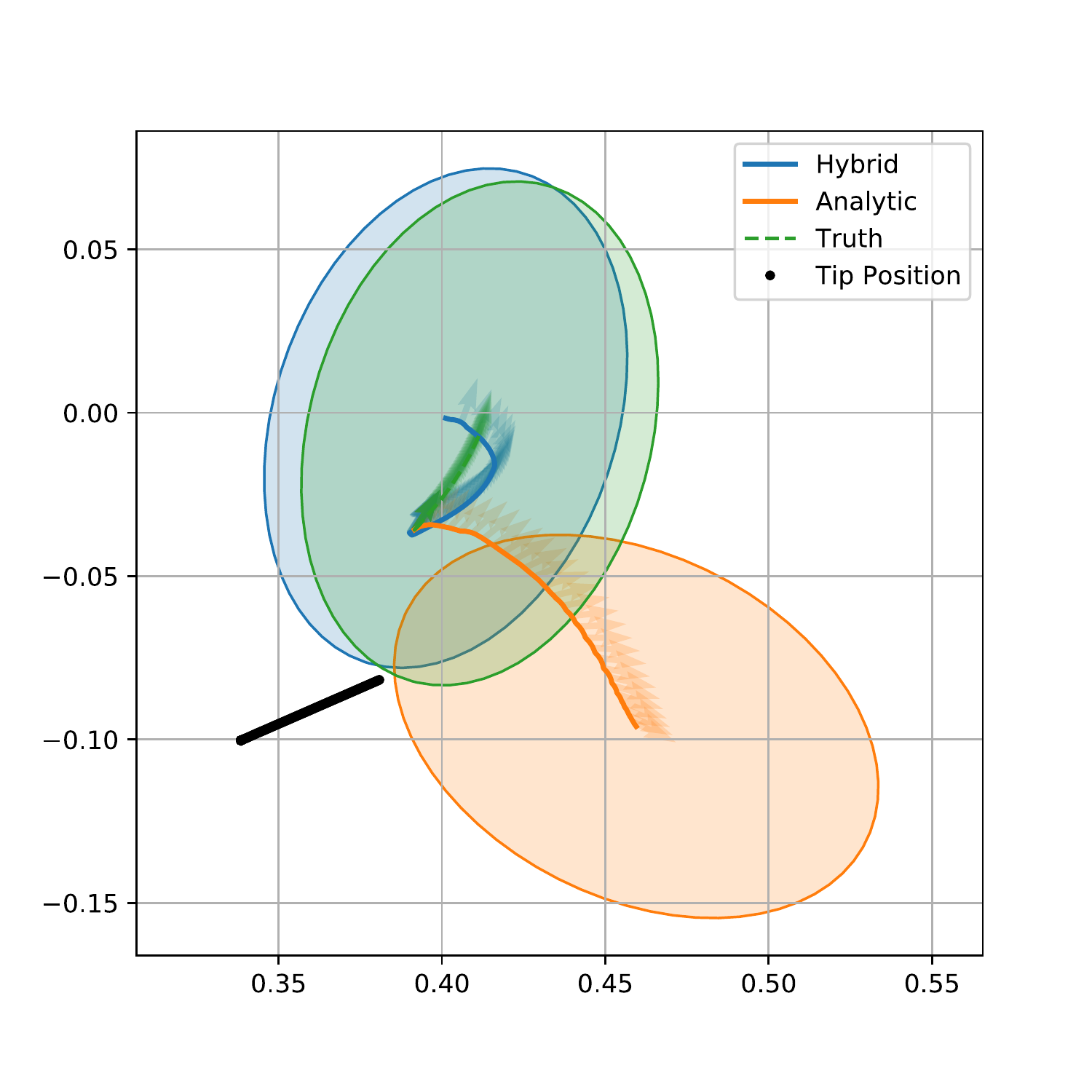}
    \caption{Trajectories of pushing an ellipsoid object from our hybrid simulator (blue) and the non-augmented rigid-body simulation (orange) compared against the real-world ground truth from the MCube push dataset~\cite{yu2016more}. The shaded ellipsoids indicate the final object poses.}
    \label{fig:push-rollout}
    \vspace*{-1em}
\end{figure}

As shown in \autoref{fig:push-butter-plywood} (left), even when these two contact models and their analytical parameters (e.g. friction coefficient, Baumgarte stabilization gains, etc.) have been tuned over a variety of demonstrated trajectories, a significant error between the ground truth and simulated pose of the object remains. By augmenting the lateral 2D friction force of the object being pushed, we achieve a significantly closer match to the real-world ground truth (cf. \autoref{fig:push-rollout}). The neural network inputs are given by the object's current pose, its velocity, as well as the normal and the penetration depth at the point of contact between the tip and the object. Such contact-related variables are influenced by the friction force correction that the neural network provides. Unique to our differentiable simulation approach is the ability to compute gradients for the neural network weights from the trajectories of the object poses compared against the ground truth, since the influence of the friction force on the object pose becomes apparent only after the forward dynamics equations have been applied and integrated.

% As baseline, we train two neural network models, a feed-forward neural network (FFN) and a long sort-term memory (LSTM). The models are trained by predicting the object's pose and velocity at the next step given its current pose, velocity, as well as the current tip position and the force measured at the tip. The FFN is a three-layer (12 hidden units) fully-connected network with ELU activations. The LSTM model has a cell with 128 hidden units connected to a linear output layer. Both models are trained with the Adam optimizer on a single NVIDIA 2080 GPU for 500 epochs. 

% \begin{figure}
%     \centering
%     \newcommand{\shapefigwidth}{2.5cm}
%     \includegraphics[width=\shapefigwidth]{fig/shape_plots/butter.pdf}
%     \includegraphics[width=\shapefigwidth]{fig/shape_plots/ellip1.pdf}
%     % \includegraphics[width=\shapefigwidth]{fig/shape_plots/ellip3.pdf}
%     \includegraphics[width=\shapefigwidth]{fig/shape_plots/hex.pdf}
%     % \includegraphics[width=\shapefigwidth]{fig/shape_plots/rect1.pdf}
%     \caption{Selection of shapes used from the MIT Push Dataset. The red dots indicate the contact points that are predefined from a grid with a resolution of \SI{2}{\cm}.}
%     \label{fig:push-shapes}
% \end{figure}
%===============================================================================

\subsection{Passive Frictional Forces for Articulated Systems}
\label{sec:exp-swimmer}

\begin{figure}
    \centering
    \includegraphics[width=0.65\columnwidth,trim={0cm -1cm 1cm 0},clip]{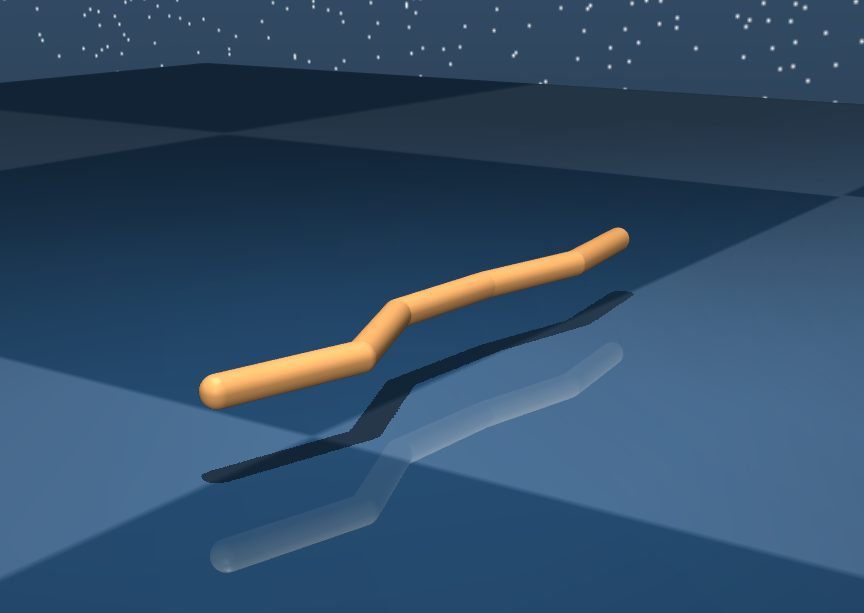}\hfill
    \includegraphics[width=0.33\columnwidth]{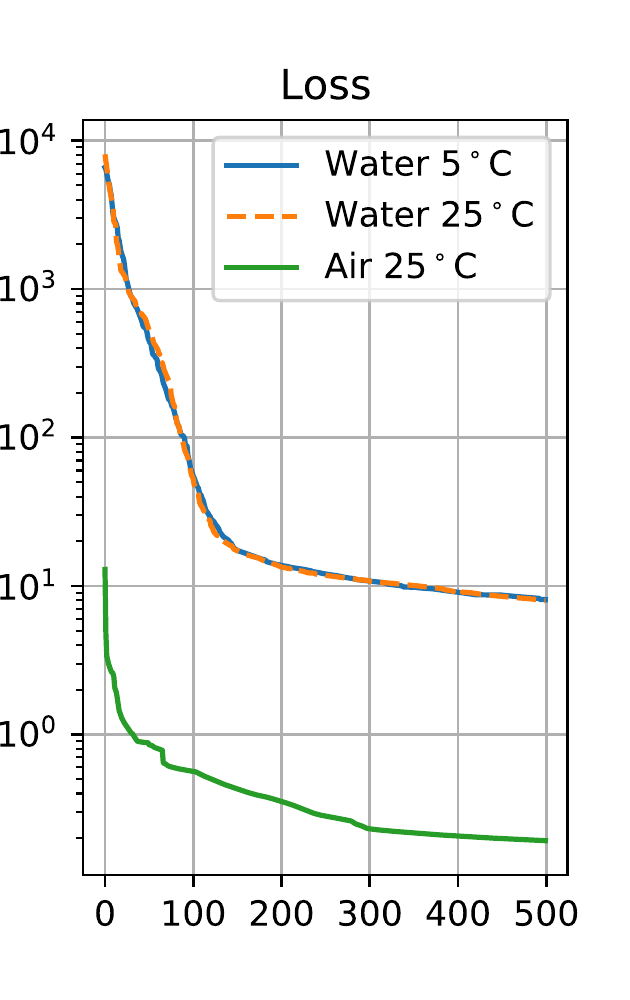}
    \caption{\textit{Left:} Model of a swimmer in MuJoCo with five revolute joints connecting its links simulated as capsules interacting with ambient fluid.
    \textit{Right:} Evolution of the cost function during the training of the neural-augmented simulator for the swimmer in different media.}
    \label{fig:swimmer05-render}
    \vspace*{-1em}
\end{figure}

\begin{figure*}
    \centering
    \includegraphics[width=\textwidth]{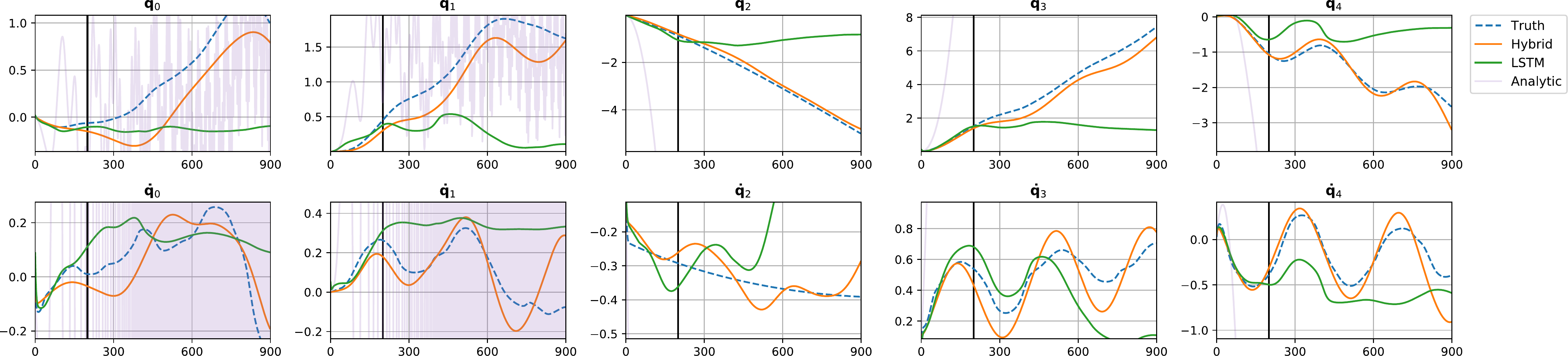}
    \caption{Trajectory of positions $\mathbf{q}$ and velocities $\mathbf{\dot{q}}$ of a swimmer in water at 25 ${}^\circ$C temperature actuated by random control inputs. In a low-data regime, our approach (orange) generates accurate predictions, even past the training cutoff (vertical black bar), while an entirely data-driven model (green) regresses to the mean. The behavior of the unaugmented analytical model, which corresponds to a swimmer in vacuum, is shown for comparison.}
    \label{fig:swimmer02-trajectory}
    \vspace*{-1em}
\end{figure*}

\begin{table}[]
    \centering
    \begin{tabular}{lSS}
    \toprule
    \bf Medium & \bf Dynamic Viscosity $\mu$ [Pa$\cdot$s] & \bf Density $\rho$ [kg / m${}^3$] \\\midrule
        Vacuum & 0 & 0 \\
        % https://onlinelibrary.wiley.com/doi/pdf/10.1002/9781118131473.app3
        Water (5 ${}^\circ$C) & 0.001518 & 1000 \\ 
        Water (25 ${}^\circ$C) & 0.00089 & 997 \\
        % https://www.engineersedge.com/physics/viscosity_of_air_dynamic_and_kinematic_14483.htm
        Air (25 ${}^\circ$C) & 0.00001849 & 1.184 \\\bottomrule\\
    \end{tabular}
    \vspace*{-1em}
    \caption{Physical properties of media used in the swimmer experiment.}
    \label{tab:media}
\vspace{-1em}
\end{table}

While in the previous experiment an existing contact model has been enriched by neural networks to compute more realistic friction forces, in this experiment we investigate how such augmentation can learn entirely new phenomena which were not accounted for in the analytical model. Given simulated trajectories of a multi-link robot swimming through an unidentified viscous medium (\autoref{fig:swimmer05-render}), the augmentation is supposed to learn a model for the passive forces that describe the effects of viscous friction and damping encountered in the fluid. Since the passive forces exerted by a viscous medium are dependent on the angle of attack of each link, as well as the velocity, we add a neural augmentation to $\mathbf{\tau}$, $\mathbf{\psi}_\theta(\mathbf{q}, \mathbf{\dot{q}})$, analogous to \autoref{fig:golf}.

Using our hybrid physics engine, and given a set of generalized control forces $u$, we integrate \autoref{eqn:augmented-manipulator-equation} to form a trajectory $\hat{T} = \{\mathbf{\hat{q}}_i, \mathbf{\hat{\dot{q}}}_i\}_i$, and compute an MSE loss $\mathcal{L}(T, \hat{T})$ against $T$, a ground truth trajectory of generalized states and velocities recorded using the MuJoCo~\cite{todorov2012mujoco} simulator. We are able to compute exact gradients $\nabla_\theta \mathcal{L}$ end-to-end through the entire trajectory, given the differentiable nature of our simulator, integrator, and function approximator.

We train the hybrid simulator under different ground-truth inputs. We consider water at 5 ${}^\circ$C and 25 ${}^\circ$C temperature, plus air at 25 ${}^\circ$C, as media, which have different properties for dynamic viscosity and density (see \autoref{tab:media}).

Leveraging the mechanical structure of problems aids in long-term physically relevant predictive power, especially in the extremely sparse data regimes common in learning for robotics. As shown in \autoref{fig:swimmer02-trajectory}, our approach, trained end-to-end on the initial 200 timesteps of 10 trajectories of a 2-link swimmer moving in water at 25 ${}^\circ$C, exhibits more accurate long-term rollout prediction over 900 timesteps than a deep long short-term memory (LSTM) network. In this experiment, the augmentation networks for TDS have 1637 trainable parameters, while the LSTM has 1900.

\subsection{Discovering Neural Augmentations}
\label{sec:discovery}

\begin{figure}[t]
    \centering
    \includegraphics[width=0.3\columnwidth]{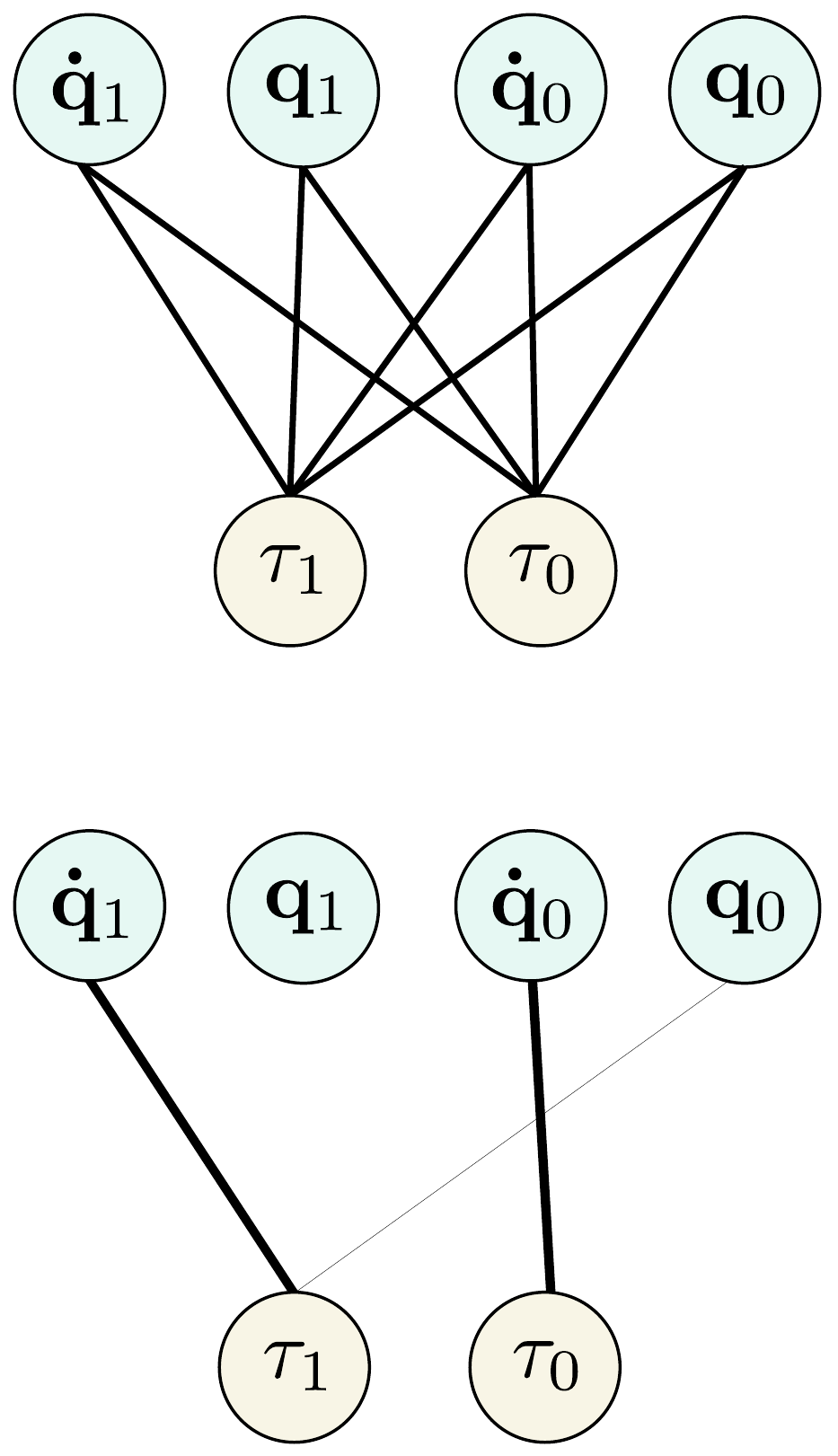}\hfill
    \includegraphics[width=0.68\columnwidth,trim=0 0 1.5cm 0]{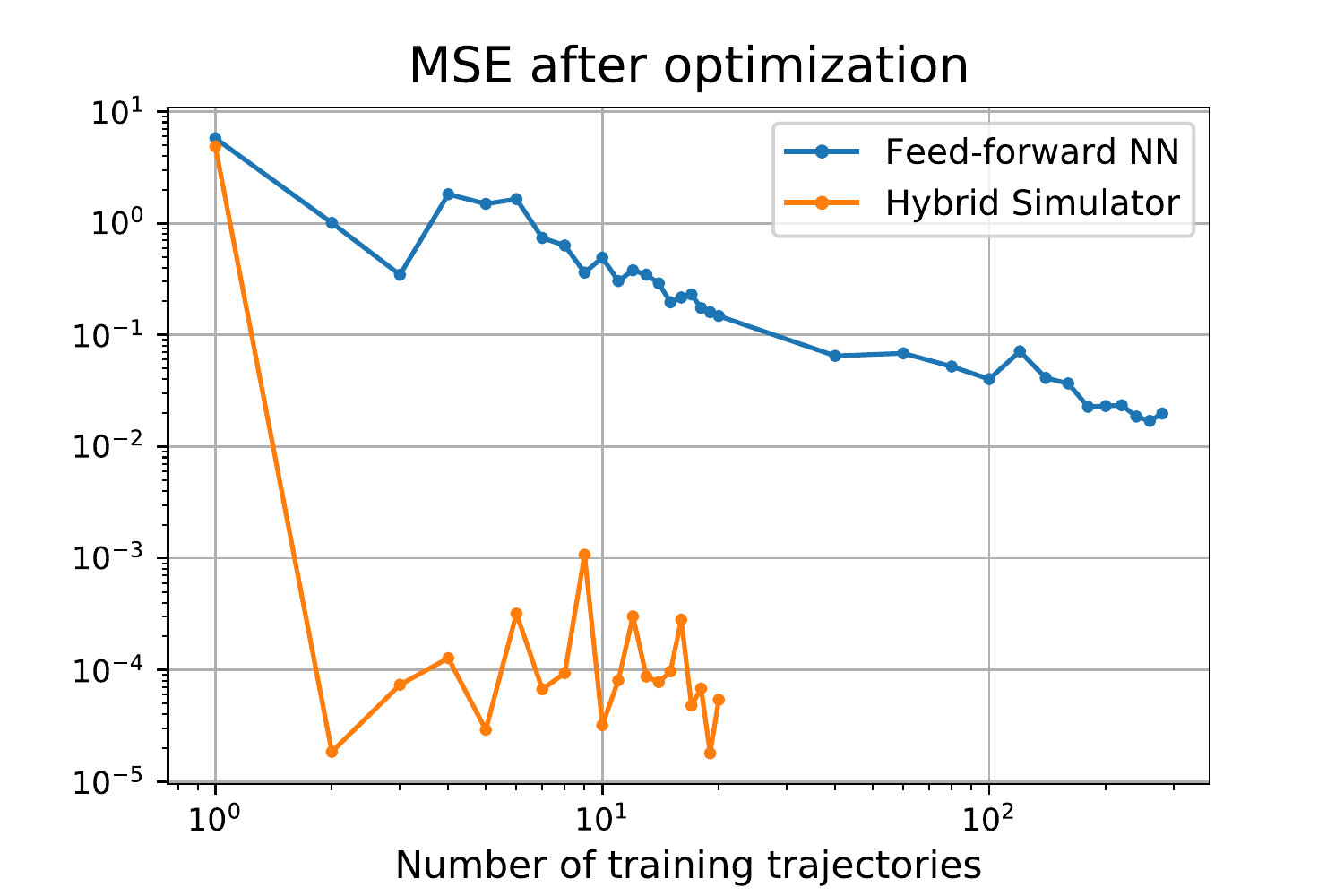}
    \caption{\textit{Left:} Neural augmentation for a double pendulum that learns joint damping. Given the fully connected neural network (top), after 15 optimization steps using the sparse group lasso cost function (\autoref{eq:spinn}) (bottom), the input layer (light green) is sparsified to only the relevant quantities that influence the dynamics of the observed system.  \textit{Right:} Final mean squared error (MSE) on test data after training a feed-forward neural network and our hybrid physics engine.}
    \label{fig:pendulum}
    \vspace*{-1em}
\end{figure}

In our previous experiments we carefully selected which simulation quantities to augment by neural networks. While it is expected that air resistance and friction result in passive forces that need to be applied to the simulated mechanism, it is often difficult to foresee which dynamical quantities actually influence the dynamics of the observed system. Since we aim to find minimally invasive data-driven models, we focus on neural network augmentations that have the least possible effect while achieving the goal of high predictive accuracy and generalizability. Inspired by Sparse Input Neural Networks (SPINN)~\cite{feng2019spinn}, we adapt the sparse-group lasso~\cite{simon2013sparse} penalty for the cost function:
\begin{align}
    \label{eq:spinn}
    \mathcal{L} = \sum_t ||f_\theta(s_{t-1}) - s^*_t||_2^2 + \kappa ||\theta_{[:1]}||_1 + \lambda ||\theta_{[1:]}||_2^2,
\end{align}
given weighting coefficients $\kappa,\lambda>0$, where $f_\theta(s_{t-1})$ is the state predicted by the hybrid simulator given the previous simulation state $s_{t-1}$ (consisting of $\mathbf{q},\mathbf{\dot{q}}$), $s^*_t$ is the state from the observed system, $\theta_{[:1]}$ are the network weights of the first layer, and $\theta_{[1:]}$ are the weights of the upper layers. Such cost function sparsifies the input weights so that the augmentation depends on only a small subset of physical quantities, which is desirable when simpler models are to be found that potentially overfit less to the training data while maintaining high accuracy (Occam's razor). In addition, the $L2$ loss on the weights from the upper network layers penalizes the overall contribution of the neural augmentation to the predicted dynamics since we only want to augment the simulation where the analytical model is unable to match the real system.

We demonstrate the approach on a double pendulum system that is modeled as a frictionless mechanism in our analytical physics engine. The reference system, on the other hand, is assumed to exhibit joint friction, i.e., a force proportional to and opposed to the joint velocity. As expected, compared to a fully learned dynamics model, our hybrid simulator outperforms the baselines significantly (\autoref{fig:pendulum}). As shown on the right, convergence is achieved while requiring orders of magnitude less training samples.
At the same time, the resulting neural network augmentation converges to weights that clearly show that the joint velocity $\mathbf{\dot{q}}_i$ influences the residual joint force $\tau_i$ that explains the observed effects of joint friction in joint $i$ (\autoref{fig:pendulum} left).

\subsection{Imitation Learning from MPC}
\label{sec:laikago}

Starting from a rigid-body simulation and augmenting it by neural networks has shown significant improvements in closing the sim-to-real gap and learning entirely new effects, such as viscous friction that has not been modeled analytically.
In our final experiment, we investigate how the use of learned models can benefit a control pipeline for a quadruped robot.
We simulate a Laikago quadruped robot and use a model-predictive controller (MPC) to generate walking gaits. The MPC implements~\cite{kim2019highly} and uses a quadratic programming (QP) solver leveraging the OSQP library~\cite{osqp} that computes the desired contact forces at the feet, given the current measured state of the robot (center of mass (COM), linear and angular velocity, COM estimated position (roll, pitch and yaw), foot contact state as well as the desired COM position and velocity). The swing feet are controlled using PD position control, while the stance feet are torque-controlled using the QP solver to satisfy the motion constraints. This controller has been open-sourced as part of the quadruped animal motion imitation project~\cite{RoboImitationPeng20}\footnote{The implementation of the QP-based MPC is available open-source at \url{https://github.com/google-research/motion_imitation}}.

We train a neural network policy to imitate the QP solver, taking the same system state as inputs, and predicting the contact forces at the feet which are provided to the IK solver and PD controller.
As training dataset we recorded 10k state transitions generated by the QP solver, using random linear and angular MPC control targets of the motion of the torso. We choose a fully connected neural network with 41-dimensional input and two hidden layers of 192 nodes that outputs the 12 joint torques, all using ELU activation~\cite{clevert2016fast}. 
% The network is trained using the Adam optimizer in 30 minutes on a single NVIDIA RTX 2080 GPU.

The neural network controller is able to produce extensive walking gaits in simulation for the Laikago robot, and in-place trotting gaits on a real Unitree A1 quadruped (see \autoref{fig:neural-qp}). The learned QP solver is an order of magnitude faster than the original QP solver (\SI{0.2}{\ms} inference time versus \SI{2}{\ms} on an AMD 3090 Ryzen CPU).

Thanks to improved performance, the control frequency increased from \SI{160}{\Hz} to \SI{300}{\Hz}. Since the model was trained using data only acquired in simulation, there is some sim-to-real gap, causing the robot to trot with a slightly lower torso height. In future work, we plan to leverage our end-to-end differentiable simulator to fine-tune the policy while closing the simulation-to-reality gap from real-world sensor measurements. Furthermore, the use of differentiable simulators, such as in~\cite{um2020sol}, to train data-driven models with faster inference times is an exciting direction to speed up conventional physics engines.

%===============================================================================

\section{Conclusion}
\label{sec:conclusion}

We presented a differentiable simulator for articulated rigid-body dynamics that allows the insertion of neural networks at any point in the computation graph. Such neural augmentation can learn dynamical effects that the analytical model does not account for, while requiring significantly less training data from the real system. Our experiments demonstrate benefits for sim-to-real transfer, for example in planar pushing where, contrary to the given model, the friction force depends not only on the objects velocity but also on its position on the surface. Effects that have not been present altogether in the rigid-body simulator, such as drag forces due to fluid viscosity, can also be learned from demonstration. Throughout our experiments, the efficient computation of gradients that our implemented simulator facilitates, has considerably sped up the training process. Furthermore, replacing conventional components, such as the QP-based controller in our quadruped locomotion experiments, has resulted in an order of magnitude computation speed-up.

In future work we plan to investigate how physically meaningful constraints, such as the law of conservation of energy, can be incorporated into the training of the neural networks. We will further our preliminary investigation of discovering neural network connections and the viability of such techniques on real-world systems, where hybrid simulation can play a key role in model-based control.

% Through gradient-based optimization, nonlinear effects can be learned by these data-driven models while their contributions are kept at a minimum through the sparse-group lasso penalty. 

% We plan to investigate applications in robotics, where the sim-to-real gap can be reduced by leveraging such hybrid simulation approach, and scale it to more complex systems involving real-world sensor measurements as training data.

% NN benefits:

% model any kind of unknown dynamics

% speed improvement over conventional algorithms (control + modeling)

% open problems:

% how to include energy constraints (i.e., only energy-conserving or at least not energy-increasing augmentations)
% add analytical gradients for LCP contact like Belbute-Perez et al.

%===============================================================================
\section*{Acknowledgments}
We thank Carolina Parada and Ken Caluwaerts for their helpful feedback and suggestions.

\bibliographystyle{IEEEtran}
\bibliography{literature}  % .bib

\end{document}